\documentclass[runningheads]{llncs}
\usepackage{graphicx}
\usepackage{hyperref}
\usepackage{balance}
\usepackage{multirow}

\begin{document}
\title{Pollen Grain Microscopic Image Classification Using an Ensemble of Fine-Tuned Deep Convolutional Neural Networks
	\thanks{This research has received funding from the Austrian Research Promotion Agency (FFG), No. 872636. We thank Nvidia corporation for their generous GPU donation.}}
%
%
\author{Amirreza Mahbod\inst{1} \and
Gerald Schaefer\inst{2} \and
Rupert Ecker\inst{3} \and
Isabella Ellinger\inst{1}}
\authorrunning{A. Mahbod et al.}
%
\institute{Institute for Pathophysiology and Allergy Research, Medical University of Vienna, Vienna, Austria  \and
Department of Computer Science, Loughborough University, Loughborough, United Kingdom \and
Research and Development Department, TissueGnostics GmbH, Vienna, Austria
}
\maketitle              
\begin{abstract}
Pollen grain micrograph classification has multiple applications in medicine and biology. Automatic pollen grain image classification can alleviate the problems of manual categorisation such as subjectivity and time constraints. While a number of computer-based methods have been introduced in the literature to perform this task, classification performance needs to be improved for these methods to be useful in practice.

In this paper, we present an ensemble approach for pollen grain microscopic image classification into four categories: Corylus Avellana well-developed pollen grain, Corylus Avellana anomalous pollen grain, Alnus well-developed pollen grain, and non-pollen (debris) instances. In our approach, we develop a classification strategy that is based on fusion of four state-of-the-art fine-tuned convolutional neural networks, namely EfficientNetB0,  EfficientNetB1, EfficientNetB2 and SeResNeXt-50 deep models. These models are trained with images of three fixed sizes ($224 \times 224$, $240 \times 240$, and $260 \times 260$ pixels) and their prediction probability vectors are then fused in an ensemble method to form a final classification vector for a given pollen grain image.

Our proposed method is shown to yield excellent classification performance, obtaining an accuracy of of 94.48\% and a weighted F1-score  of 94.54\% on the ICPR 2020 Pollen Grain Classification Challenge training dataset based on five-fold cross-validation. Evaluated on the test set of the challenge, our approach achieved a very competitive performance in comparison to the top ranked approaches with an accuracy and a weighted F1-score of 96.28\% and 96.30\%, respectively. 

\end{abstract}

\section{Introduction}
Palynology is the scientific study of pollen grains which are produced by plants for the purpose of reproduction. Nowadays, pollen grain classification is a valuable tool for various applied sciences including systematics and forensics~\cite{Halbritter2018}. In the medical context, pollens are among the most common triggers of seasonal allergies. Allergic diseases have become a major public health problem. The World Health Organization (WHO) estimates that worldwide 400 million people suffer from allergic rhinitis~\cite{cruz2007global}. In Europe, the prevalence of pollen allergy in the general population is about 40\%~\cite{doi:10.1111/j.1398-9995.2007.01393.x}. Clinicians and patients may be able to anticipate the onset of pollen-related allergy symptoms by monitoring pollen levels~\cite{doi:10.1111/all.13758}.

The performance of any pollen counter (human or machine) depends on two key tasks: finding the pollen grains by discriminating pollen from non-pollen, and pollen grain classification. Traditionally, pollen grain classification involves observation and discrimination of features by a highly qualified palynologist. While this is an accurate and effective method, it takes considerable amounts of time and resources~\cite{stillman1996needs}. Therefore, automatic recognition of pollen species by means of computer vision is of major importance in modern palynology~\cite{doi:10.1111/nph.12848}.

Automatic computer-based pollen grain image classification methods have been proposed in several studies with the aim of reducing human interaction in the analysis procedure~\cite{doi:10.1111/nph.12848}. Similar to other medical or natural image classification tasks, machine learning-based approaches have shown a better classification performance for pollen grain image classification compared to standard image processing techniques~\cite{Battiato_2020_CVPR_Workshops,sevillano2018improving}.

Most of the classical machine learning-based approaches for pollen grain image classification made use of feature engineering to train a classifier such as multi-layer perceptron or support vector machines. Well-known image features that were used for this task include morphological features~\cite{5954712}, texture features~\cite{fernandez2003improved} or hybrid features~\cite{doi:10.1002/jemt.22091}. However, due to the similar morphological appearance of pollen grains and various artefacts that may present, deriving well-working hand-crafted features is a challenging task.

Using deep learning approaches and more specifically convolutional neural networks (CNNs)~\cite{litjens2017survey} allows for a more suitable solution for distinguishing different types of pollen grains compared to classical machine learning approaches. These deep models can be trained end-to-end without the use of hand-crafted features and can be applied directly on raw or pre-processed images~\cite{goncalves2016feature,sevillano2018improving,Battiato_2020_CVPR_Workshops,10.1007/978-3-319-50835-1_30}.

Since the number of labelled pollen grain images in the publicly available datasets is too small to train a CNN from scratch, transfer learning can be employed. Here, pre-trained CNNs can be used as optimised deep feature extractors or they can be fine-tuned to solve the classification task. Standard pre-trained CNNs such as AlexNet~\cite{Krizhevsky2012} or VGGNet~\cite{Simonyan2014} have been used for these purposes~\cite{Battiato_2020_CVPR_Workshops,sevillano2018improving,menad2019deep,sevillano2020precise}. However, the performance of such approaches still needs to be improved to be useful for practical applications.

In this paper, we present an ensemble method for multi-class pollen grain microscopic image classification. Based on our earlier work on various medical image classification tasks and recent advances in transfer learning~\cite{mahbod2020investigating,MAHBOD2020105475,mahbod2018breast,10.1145/3390557.3394303}, we develop a classification method which exploits four state-of-the-art pre-trained CNNs fine-tuned with pollen grain images at three different image resolutions. We then fuse the prediction vectors from different models in an ensemble strategy to form the final classification vector. Evaluated on the training set and the test set of the ICPR 2020 Pollen Grain Classification Challenge\footnote{{\url{https://iplab.dmi.unict.it/pollenclassificationchallenge}}}, our proposed method provides excellent classification performance, achieving weighted F1-scores of 94.54\% and 96.30\% on the challenge training and test set, respectively.

\section{Materials and Methods}
\subsection{Datasets}
\begin{figure}[t!]
	\centering
	\includegraphics[width=.7\columnwidth]{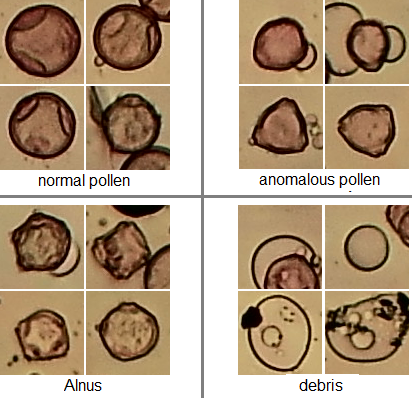} 
	\caption{Example images from the ICPR 2020 Pollen Grain Classification Challenge.}
	\label{example}
\end{figure}

While a number of earlier datasets exist for pollen grain image classification, such as the POLEN23E dataset~\cite{sevillano2018improving} or the dataset from Ranzato {\it et al.}~\cite{ranzato2007automatic}, which contain 805 and 1,429 pollen grain images, respectively, in this paper, we use the ICPR 2020 Pollen Grain Classification Challenge dataset\footnote{\url{https://iplab.dmi.unict.it/pollenclassificationchallenge/train.zip}}, which is one the biggest publicly available datasets for this task~\cite{Battiato_2020_CVPR_Workshops}. This dataset contains 11,279 microscopic training image of size $84 \times 84$ pixels in RGB format. The images belong to one of four classes, namely Corylus Avellana well-developed pollen grain (normal pollen), Corylus Avellana anomalous pollen grains (anomalous pollen), Alnus well-developed pollen grains (Alnus), and non-pollen (debris), example of which are shown in Fig.~\ref{example}. There are 1566 normal pollen images, 773 anomalous pollen images, 8216 Alnus images, and 724 debris images. In addition to the raw images, segmentation masks of the pollen grains are also provided by the challenge organisers, but we did not make use of these. A test set of 1,991 images (without segmentation masks) has also been released but labels for these images are kept private by the challenge organisers. 


\subsection{Pre-processing}
\label{preprocess}
We apply two pre-processing steps. First, we subtract the mean intensity RGB values of the ImageNet dataset~\cite{Deng2009} from all training and test images. Then we resize all images to a fixed size of $260 \times 260$ pixels. 

\subsection{Pre-trained CNNs}
There are a number of well-established pre-trained CNNs that can be employed for fine-tuning. Recently developed deep neural networks that are widely used for various transfer learning applications include but are not limited to residual networks and their derivatives such as ResNeXt or wide ResNet~\cite{He2016,8100117}, DenseNets with various depths such as Densenet-121 or DenseNet-169~\cite{huang2017densely}, squeeze and excitation networks that can be also combined with other models such as SeResNeXt or SeResNet~\cite{hu2018squeeze}, GoogLeNet and its derivatives such as Inceptions~\cite{Szegedy2015,szegedy2016rethinking} and EfficientNet~\cite{tan2019efficientnet} models (the state-of-the-art models for ImageNet classification task). As mentioned above, some earlier pre-trained CNNs such as AlexNet or VGGNet have already been used for pollen grain image classification. However, their classification performance can be improved by employing more advanced architectures which for other tasks have shown superiority over AlexNet or VGGNet. 

In our approach, we exploit four recently developed yet already well-established pre-trained CNNs from the EfficientNet~\cite{tan2019efficientnet} and SeResNeXt model~\cite{hu2018squeeze} families as they have shown excellent performance in, for example, classification of skin lesions or ophthalmological image classification~\cite{MAHBOD2020105475,10.1145/3390557.3394303,MAHBOD2020105725}.

The backbone model of the SeResNeXt network is ResNet. ResNet models consist of special building blocks, called residual blocks, that alleviate the vanishing gradient problem by connecting the input and output of each residual block. Thus, the network depth can be increased to yield a better classification performance with deeper models. ResNet-50, ResNet-101, and ResNet-152 are the three most used variations, while ResNeXt is a modified architecture whose residual blocks are wider compared to ResNet through multiple parallel pathways similar to the inception module in the GoogLeNet family. SeResNeXt incorporates the squeeze and excitation blocks from~\cite{hu2018squeeze} into the model, while SeResNeXt-50 and SeResNeXt-101 are the best-known models of the family.

In general, by solving the vanishing gradient issue by some techniques such as skip or dense connections, the performance of a CNN can be improved by increasing the network depth, width, or input image resolution. In the design of many former CNNs such as ResNet or wide ResNet, only one aspect of the model is increased which leads to a better classification performance (e.g. DenseNet-169 has a better classification score in comparison to DenseNet-121 as it is deeper). In contrast, in the EfficientNet architecture, all three aspects of the models are increased systematically and by a constant factor. This leads to generation of eight different architectures, EfficientNetB0 to EfficientNetB7, that have various depths, widths and pre-defined input image sizes (from $224 \times 224$ to $600 \times 600$ pixels) while being computationally less expensive compared to other well-known CNNs.    

We use three variations of EfficientNet, namely EfficientNetB0, EfficientNetB1, and EfficientNetB2, as well as the SeResNeXt-50 model for fine-tuning. We choose the shallower versions of the EfficientNet family networks to prevent over-fitting to the limited training data available.  All four utilised networks are initially trained on natural images with image sizes between $224 \times 224$ to $260 \times 260$ pixels

\subsection{Fine-tuning}
\label{fine-tuning}
\begin{figure}[t!]
	\centering
	\includegraphics[width=\columnwidth]{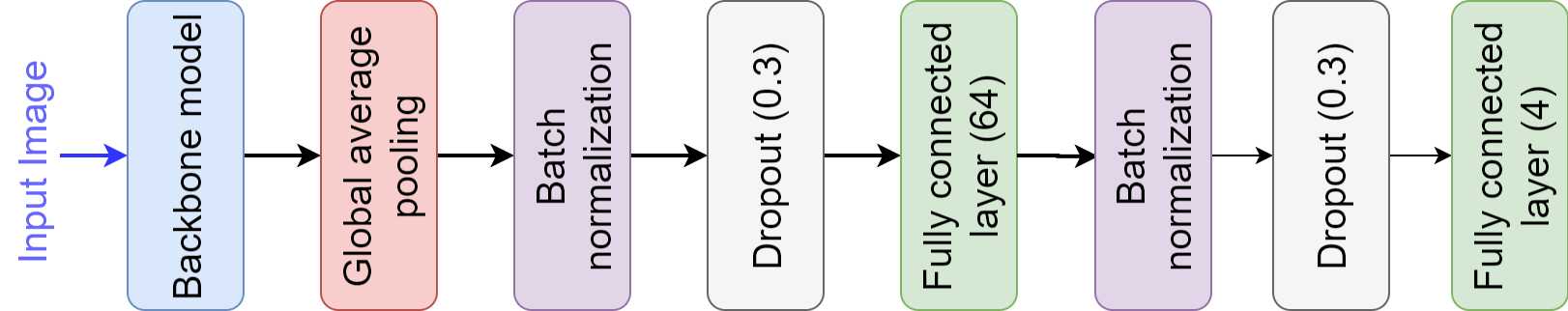} 
	\caption{Graphic representation of the fine-tuning approach. The backbone model is either EfficientNetB0, EfficientNetB1, EfficientNetB2 or SeResNeXt-50.}
	\label{fine-tuning_fig}
\end{figure}

We apply a similar approach for fine-tuning the networks as in our work on skin lesion classification~\cite{MAHBOD2020105475,mahbod2020investigating,mahbod2019fusing}. The general scheme for fine-tuning the pre-trained networks is depicted in Fig.~\ref{fine-tuning_fig}.

\begin{figure}[t!]
	\centering
	\includegraphics[width=0.8\columnwidth]{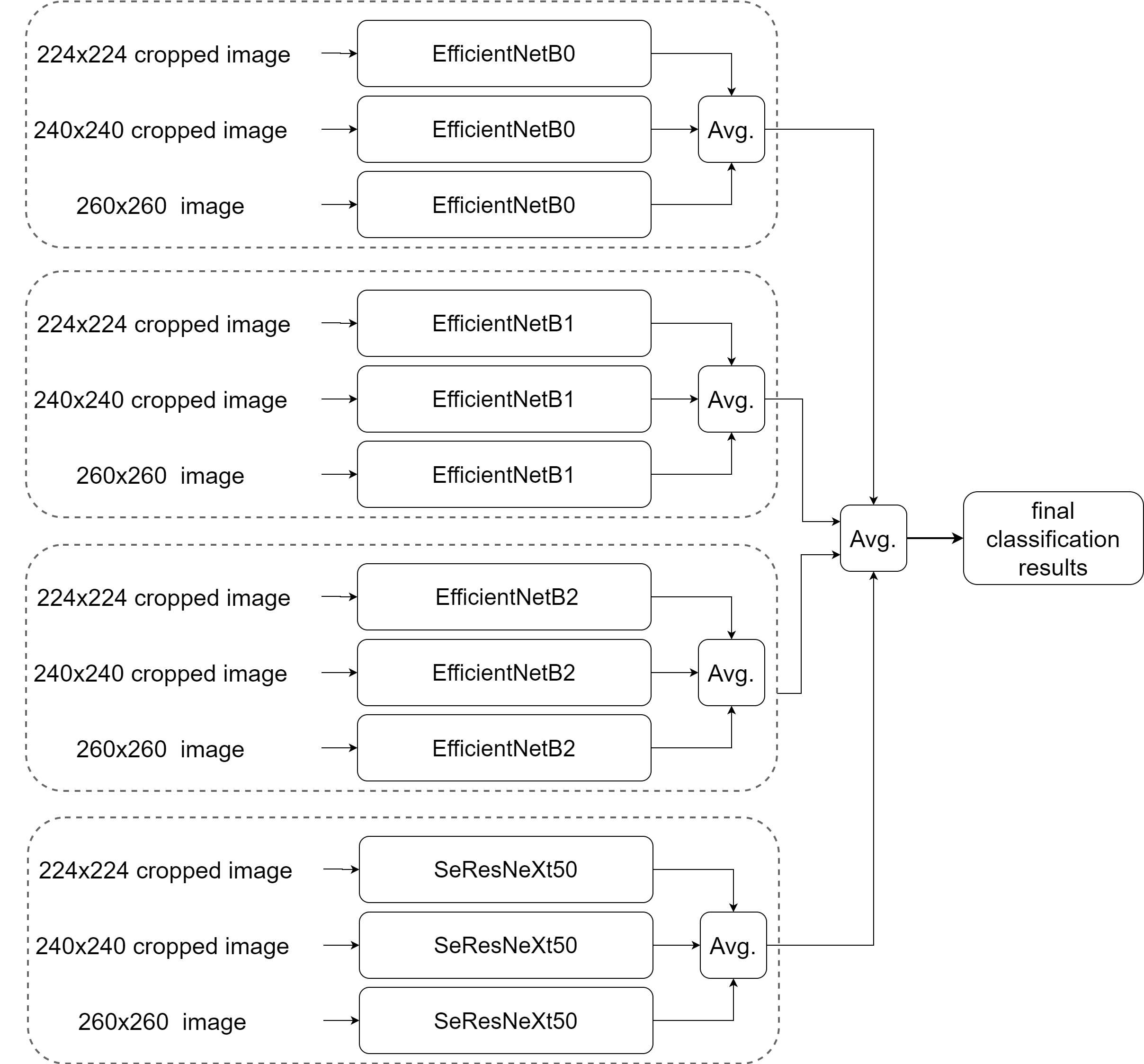} 
	\caption{Graphic representation of of the proposed ensemble strategy.}
	\label{ensemble}
\end{figure}

To fine-tune the pre-trained networks, first fully connected (FC) layers of the original networks are removed. Then, we apply two blocks of batch normalisation, drop out layers (with a drop factor of 0.3) and FC layers (with 64 and 4 nodes in the first block and second block, respectively). We train the models with three different images sizes, namely $224 \times 224$,  $240 \times 240$, and $260 \times 260$ pixels. For the two smaller sizes (i.e. $224 \times 224$ and $240 \times 240$ pixel images), random cropping within the $260 \times 260$ pixel images is performed during training and testing.  We use a global average pooling layer just before the first batch normalisation layer to prevent dimensionality mismatch for different image sizes.

The weights of the newly added FC layers are initialised with the Xavier method~\cite{glorot2010understanding} and the weight factors are kept 10 times larger compared to all other learnable weights in the models. We fine-tune the network with the Adam optimiser~\cite{Kingma2014} with an initial learning rate of 0.001. We halve the learning rate after every seven epochs and train the networks in total for 40 epochs. To deal with the imbalanced dataset set, we use a weighted focal loss function~\cite{lin2017focal} with weights of 7.20, 14.59, 1.37, and 15.57 for the normal pollen, anomalous pollen, Alnus pollen, and debris, respectively.

As in~\cite{MAHBOD2020105475}, we use various augmentation techniques in the training and testing phases to prevent over-fitting. For training augmentation, we use random cropping (for the models trained with $224 \times 224$ and $240 \times 240$ pixels images), adaptive histogram equalisation (with a probability of 0.1), random cut-out (with a probability of 0.1), random brightness and contrast shifts (with a probability of 0.4), random horizontal and vertical flipping (with a probability of 0.5), and random rotations (0, 90, 180 and 270 degrees, with a probability of 0.5).

In the inference phase, we use 25-folds test time augmentation with the same augmentation methods as described for training. We monitor the prediction probability vectors for each augmented image. If the maximum value in a prediction vector is below 0.5, we disregard that prediction vector and take the average over the rest of the prediction vectors for a specific test image.

\subsection{Fusion}
To improve the classification performance, we employ a straightforward ensemble strategy as shown in Fig.~\ref{ensemble}.

In our fusion scheme, first, the results from the five folds for each image size and for each network are fused. Then, the results from three different image sizes for a specific model architecture are combined and finally the classification probability vectors of the four utilised model architectures (i.e. EfficientNetB0, EfficientNetB0, EfficientNetB0, and SeResNeXt-50) are fused to form the final classification predictions. From the final prediction for each image, we chose the element with maximum probability to determine the image class. Fusion of multiple models is performed by taking the average over their classification prediction vectors.

\subsection{Evaluation}
As evaluation measures, we use accuracy, balanced accuracy~\cite{10.1109/ICPR.2010.764}, which is defined as the average recall over all classes, and the weighted F1 score~\cite{chinchor1995message}, which is calculated as the F1 score for each class weighted by its number of true samples. The latter is also the main evaluation measure that is used in the ICPR challenge.

\subsection{Implementation}
We use Keras\footnote{\url{https://keras.io/}} and Tensorflow\footnote{\url{https://www.tensorflow.org/}} deep learning frameworks for algorithm development. All experiments are conducted using a single workstation with an Intel Core i7-8700 3.20 GHz CPU, 32 GB of RAM and a Titian V Nvidia GPU card with 12 GB of installed memory.

\section{Results}
\label{result}
Since the ground truth of the test data is kept private by the challenge organisers, in developing our model and for showing the effectiveness of the utilised ensemble approach, we use five-fold cross-validation (5CV) based on the ICPR challenge training data. That is, the training data is split into five folds and four of these are used for training while the fifth is utilised for testing. This is repeated five times so that each test fold is used once and the obtained results are averaged. 


Table~\ref{size} shows the 5CV results for each network architecture and each input image size. 

\begin{table}[t!]
	\centering
	\caption{5CV results [\%] based on different input crop sizes and different network architectures. }
	\label{size}
		\makebox[\linewidth][c]{
	\begin{tabular}{lcccc}
		\hline
		\hline
		\textbf{network}  & \textbf{image size} & \textbf{accuracy} & \textbf{balanced accuracy} & \textbf{weighted F1-score}\\
		\hline
		\hline
		EfficientNetB0            & $224\times224$  & 92.48 & 92.20 & 92.66\\
		EfficientNetB0            & $240\times240$  & 91.96 & 92.22 & 92.18  \\
		EfficientNetB0            & $260\times260$  & 93.07 & 92.18 & 93.20 \\
		\hline
		EfficientNetB1            & $224\times224$  & 93.01 & 92.11 & 93.16 \\
		EfficientNetB1            & $240\times240$  & 94.06 & 92.68 & 94.13   \\
		EfficientNetB1            & $260\times260$  & 93.20 & 92.60 & 93.34  \\
		\hline
		EfficientNetB2            & $224\times224$  & 92.74 & 92.03 & 92.91  \\
		EfficientNetB2            & $240\times240$  & 92.68 & 92.38 & 92.84  \\
		EfficientNetB2            & $260\times260$  & 93.18 & 92.17 & 93.31 \\
		\hline
		SeResNeXt-50              & $224\times224$  & 92.00 & 91.84 & 92.20  \\
		SeResNeXt-50              & $240\times240$  & 92.40 & 91.81 & 92.57  \\
		SeResNeXt-50              & $260\times260$  & 91.50 & 91.49 & 91.74  \\
		\hline
		\hline
	\end{tabular}
	}
\end{table}

The results in Table~\ref{level2} give the performance of the classification models obtained by fusing networks trained at the three different image sizes. The final row in Table~\ref{level2} shows the final fusion performance, derived from fusion of 12 (3 image sizes $\times$ 4 architectures) sub-models. 

\begin{table*}[t!]
	\centering
	\caption{5CV results [\%] based on network fusion with different input image sizes and fusion of all architectures.}
	\label{level2}
	\begin{tabular}{lcccc}
		\hline
		\hline
		\textbf{network}  & \textbf{image size} & \textbf{accuracy} & \textbf{balanced accuracy} & \textbf{weighted F1-score}\\
		\hline
		\hline
		EfficientNetB0            & all sizes  & 93.35 & 92.75  & 93.47  \\
		EfficientNetB1            & all sizes  & 94.26 & 92.83  & 94.34\\
		EfficientNetB2            & all sizes  & 93.61 & 92.52  & 93.72\\
		SeResNeXt-50              & all sizes  & 93.45 & 92.38  & 93.56 \\
		\hline
		all networks              & all sizes  & 94.48 & 93.22 &  94.54\\
		\hline
		\hline
	\end{tabular}
\end{table*}

We also compare the obtained results with other state-of-the-art classification models for pollen grain images. In particular, we compare with the results reported in~\cite{Battiato_2020_CVPR_Workshops} where both classical feature extraction-based methods and deep learning models were evaluated. The investigated techniques include histogram of oriented gradients (HOG)~\cite{1467360} and local binary pattern (LBP)~\cite{1017623} features, and multi-layer perceptron (MLP)~\cite{gardner1998artificial} and support vector machine (SVM)~\cite{cortes1995support} classifiers, while the explored deep learning models include AlexNet~\cite{Krizhevsky2012} and VGG~\cite{Simonyan2014}.

The results are listed in Table~\ref{comparion}. It should be noted that the results in~\cite{Battiato_2020_CVPR_Workshops} are obtained based on a fixed split into 85\% training data and 15\% test data and that this split is not publicly available. In contrast, we perform 5CV to evaluate our method, and thus use slightly less training data while ensuring that all available data is used once for testing.

\begin{table*}[t!]
	\centering
	\caption{Comparison to other state-of-the-art methods from~\cite{Battiato_2020_CVPR_Workshops} based on the ICPR 2020 Pollen Grain Classification Challenge training data.}
	\label{comparion}
	\begin{tabular}{llccc}
		\hline
		\hline
		\textbf{approach type}  &\textbf{features/model} & \textbf{accuracy} & \textbf{weighted F1-score} \\
		\hline 
		\hline
		classical machine learning & HOG features + RBF SVM  & 86.58 &  85.66 \\
		classical machine learning & HOG features + MLP & 84.93 & 84.31 \\
		classical machine learning & LBP features + MLP & 80.02 &  77.64  \\
		\hline		
		deep learning & AlexNet + augmentation  & 89.63 & 88.97  \\
		deep learning & small VGG + augmentation & 89.73&  89.14  \\
		\hline
		deep learning & our approach  & 94.48 &  94.54  \\
		\hline 
		\hline
	\end{tabular}
\end{table*} 

\begin{figure}[t!]
	\centering
	\begin{tabular}{ccc}
		\includegraphics[width=2.5cm]{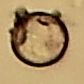} &
		\includegraphics[width=2.5cm]{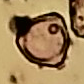} &
		\includegraphics[width=2.5cm]{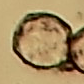} \\
		normal pollen & anomalous pollen & Alnus \\
		\includegraphics[width=2.5cm]{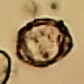} & 
		\includegraphics[width=2.5cm]{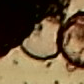} &
		\includegraphics[width=2.5cm]{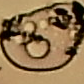} \\
		Alnus & debris & debris \\
	\end{tabular}
	\caption{Examples of correctly classified images.}
	\label{correct}
\end{figure}

\begin{figure}[b!]
	\centering
	\begin{tabular}{cc}
		\includegraphics[width=2.5cm]{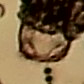} &
		\includegraphics[width=2.5cm]{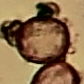} \\
		normal pollen (debris) & anomalous pollen (Alnus) \\
		\includegraphics[width=2.5cm]{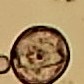} &
		\includegraphics[width=2.5cm]{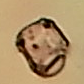} \\
		Alnus (normal pollen) & debris (Alnus)
	\end{tabular}
	\caption{Examples of incorrectly classified images. The ground truth class is mentioned first and the predicted class in (brackets).}
	\label{incorrect}
\end{figure}

Examples of correctly or incorrectly classified images in the training data (by the full fusion approach) are shown in  Fig.~\ref{correct} and Fig.~\ref{incorrect}, respectively.

As the final experiment, we extend our fusion approach to combine the output of all 5CV results in the ensemble strategy to investigate the performance of the model for the 1,991 test images of the ICPR challenge. Hence, instead of combining 12 sub-models, we fuse the results of all 60 sub-models (5CV $\times$ 3 image sizes $\times$ 4 architectures). The results for this (evaluated by the challenge organisers) and comparison to the top three performers of the ICPR challenge for the test data are shown in Table~\ref{test_res}. 

\begin{table*}[t!]
	\centering
	\caption{Results on ICPR Challenge test data of our extended fusion approach and the top three performers of the challenge. }
	\label{test_res}
	\begin{tabular}{lccc}
		\hline
		\hline
		\textbf{User name}   & \textbf{accuracy} & \textbf{weighted F1-score} \\
		\hline 
		\hline
		Zhangbaochang    & 97.53 & 97.51  \\
	   Fang Chao  &97.34 &  97.30  \\
		Penghui Gui  & 97.29  & 97.26   \\
		\hline		
	
		our approach    & 96.28 &  96.30  \\
		\hline 
		\hline
	\end{tabular}
\end{table*} 

\section{Discussion}
The results in  Table~\ref{size} show that all networks deliver very good classification performance, while EfficientNetB1 is slightly superior compared to the rest and SeResNeXt-50 is slightly worse.  

Looking at Table~\ref{level2}, we can see that fusing the results from different image sizes improves the classification performance for all four models. The performance is further slightly improved when the results from different model architectures are also fused as shown in the bottom row of Table~\ref{level2}. 

The comparison between our proposed method and other state-of-the-art techniques in Table~\ref{comparion} demonstrates a clear superiority of our approach over the other classification models. While, as mentioned, the training/test configuration we employ is somewhat different from that of the other methods, it represents arguably a harder test, yet as the results in the table clearly indicate we achieve significantly better results both in terms of accuracy and weighted F1 score. In addition, the techniques reported in~\cite{Battiato_2020_CVPR_Workshops} utilise the segmentation masks (which are only available for the training dataset of the challenge) to obtain segmented images. In contrast, our approach uses only the raw images as input and is segmentationless.


Finally, the results in Table~\ref{test_res} show the performance of our proposed method on the actual test data of the ICPR challenge. By comparing the 5CV results (last row in Table~\ref{comparion}) and the test results (last row of Table~\ref{test_res}), performance improvement in both accuracy and weighted F1 score can be observed. Compared to the top performers of the ICPR challenge our approach delivers very competitive performance with slightly inferior results (by 1.25\% and 1.21\% in terms of accuracy and weighted F1 score, respectively). Further details of the test set results can be found on the challenge website\footnote{\url{https://iplab.dmi.unict.it/pollenclassificationchallenge/results}}. Note that the details of the other approaches are not disclosed at the time of writing this manuscript.

\section{Conclusions}
In this paper, we have proposed an effective approach to classify pollen grain microscopic images to four pollen grain types. We exploit several fine-tuned CNNs at various image sizes and combined them in a simple yet effective fusion framework. Our approach does not require segmentation and we have demonstrated excellent classification results on the challenge's training  and test data sets. The performance can be potentially further improved by incorporating other network architectures or other image resolutions.

\bibliographystyle{splncs04}
\bibliography{refs}
\end{document}